# Multispectral 3D mapping on a Roman sculpture to study ancient polychromy

Francesca Uccheddu[1][0000-0001-7604-5182] Umair Shafqat Malik[2][0000-0003-2942-1063]
Emanuela Massa[3] Anna Pelagotti[4][0000-0002-6582-0320] Maria Emilia Masci[5]
Gabriele Guidi[2][0000-0002-8857-0096]

[1] University of Padova, Padua, Italy
[2] Indiana University, Bloomington, IN, USA
[3] Art-Test, Firenze, Italy
[4] Istituto Nazionale di Ottica (INO), Florence, Italy
[5] Opificio delle Pietre Dure, Florence, Italy
`gabguidi@iu.edu`

**Abstract.** Research into the polychromy of Greek and Roman sculptures has surged to explore the hypothesis that ancient sculptures were originally not pristine white but adorned with colors. Multispectral and multimodal imaging techniques have been crucial in studying painted surfaces, revealing polychromies even in traces. In fact, imaging techniques, such as reflectance and fluorescence, can identify different materials and map inhomogeneities, guiding further investigations such as Raman, XRays Fluorescence, and Fourier Transform InfraRed Spectroscopy (FTIR) to investigate residual colors. However, this approach may underestimate the original polychromies' extent over the complex articulation of a sculptured surface. This study proposes a methodology to analyze the original appearance of ancient sculptures using reality-based 3D models with textures not limited to those visible to the naked eye. We employ Visible Reflected Imaging (VIS) and Ultraviolet-induced Fluorescence Imaging (UVF). From the UVF and VIS datasets, the underlying 3D model is built by means of photogrammetry. Through raw data processing, images taken with different illuminating sources are successfully aligned and processed, creating a single 3D model with multiple textures mapped onto the same bi-dimensional space. The pixel-to-pixel correspondence of different textures allows for the implementation of a classification algorithm that can directly map its outcome onto the 3D model surface. This enables conservators to deepen their understanding of artifact preservation, observe material distribution in detail, and correlate this with 3D geometrical data. In this study, we experiment with this approach on an ancient Roman sculpture of Artemis, conserved at the Archeological and Art Museum of Maremma (MAAM) in Grosseto, Italy.

# 1 Polychromy Investigation

Research into polychromy focuses on revealing and documenting the colorful embellishments of ancient sculptures and architecture, enriching our knowledge of ancient artistic practices and cultural contexts. Although white marble has long been associated with antiquity and the classical ideal [1, 2], this perception is misleading; the absence of visible color today does not accurately represent ancient visual culture. The Greeks and Romans used a wide array of colors in their artworks, making the study of polychromy vital for a complete understanding of these civilizations [3]. However, investigating Greco-Roman polychromy presents several challenges due to the delicate nature of the painted layers, which often deteriorate after prolonged exposure to the elements once unearthed. Furthermore, many ancient restorers, adhering to the neoclassical preference for white purity, meticulously cleaned these surfaces after excavation, removing much of the visible paint. As a result, the remaining pigments are scarcely present on the sculpture surface and require specific investigation technologies.

## 1.1 Previous works

Various scholars have made substantial contributions to the study of polychromy in ancient sculptures, employing a range of techniques to uncover the original appearances and materials used in ancient pigments.

For example, a systematic study was then conducted in 2010 on the "Treu Head," a Roman fragment conserved at the British Museum. The head's polychromy was extensively investigated using noninvasive techniques such as UVF and Visible-Induced Luminescence imaging (VIL) as well as analytical methods on micro samples, including Raman Spectroscopy, FTIR, High-Performance Liquid Chromatography (HPLC), and Gas Chromatography-Mass Spectrometry (GC-MS), which revealed a complex mixture of pigments [4]. Another pivotal contribution involving the Copenhagen Polychromy Network focused on the study of ancient sculptural polychromy, primarily within the collections of the Ny Carlsberg Glyptotek [5]. The report documents extensive research into the color layers and techniques used on various sculptures, where UVF and VIL imaging were conducted on Cross-Section, Microscopy, like EMP, XRF, GC-MS, providing a comprehensive overview of the methodologies and findings of this collaborative effort.

In [6] the author analyzes the use of gilding, providing also a more expanded analysis of the different types of pigments associated with different polychrome techniques and a deeper understanding of Roman polychromy.

The scientific study of pigments has brought some experts to attempt the reconstruction in various materials of the supposed original coloring of specific ancient sculptures. A systematic research project on polichromy on Greek and Roman marble and bronze statues has been carried out by Vinzenz Brinkmann and Ulrike Koch-Brinkmann [7, 8]. It produced scientific reconstructions shown in various exhibitions starting with "Gods in Color – Painted Sculpture in Classical Antiquity" in 2008, followed by "Gods in Color – Golden Edition: Polychromy in Antiquity" in 2020. At the Vatican Museums, a physical replica of the Augustus of Prima Porta was manufactured and painted according to the results of the analysis described above [9]. Later, a digital replica of the Roman sculpture of Caligula conserved at the Virginia Museum of Fine Arts in Richmond (VA), USA, was studied with similar methodologies and digitally restored and recolored using an analogous approach based on the integration of the sampled colors and the philological reconstruction of the typical imperial togas [10]. This line of work, despite originating interesting reinterpretations of well-known masterpieces of the past, has also raised some criticism. In this regard [11] critically examines the ethical and practical considerations of reconstructing polychromy, arguing for a balanced approach that respects both the historical authenticity and the need for modern interpretative reconstructions.

Our contribution intends to provide a powerful tool to help conservators and scholars identify and map materials, even when only present in traces, directly on the 3D model.

### 1.2 The Artemis from Castiglion della Pescaia

The methods presented in this paper were tested on the Roman statue of Artemis, found in 1880 near Castiglion della Pescaia (Grosseto, Italy). This sculpture is currently undergoing restoration and diagnostic investigations at the Opificio delle Pietre Dure (OPD) in Florence. It is a Roman-era statue, dated between 50 and 20 BC, made of marble—probably Greek—in an "archaizing" style. According to a study published in 2004 [12] it is one of several marble replicas derived from a bronze original of the classical period, dated between 500 and 480 BC, located in Segesta, Sicily. Taken to Rome, it became famous and was replicated in various marble copies.

Today, four marble replicas of this type of Artemis are known:
- The replica in the National Archaeological Museum of Venice, inv. 59 [13]
- The statue of Artemis found in a domus in Pompeii in 1760, now in the National Archaeological Museum of Naples (inv. 6008), which has been particularly studied for the remnants of polychromy it preserves [14].
- The one from Castiglion della Pescaia, housed in the Archaeological and Art Museum of Maremma (MAAM) in Grosseto, inv. 10701 [15, 16], now under restoration at the OPD.
- The copy seized in 2001 by the Artistic Heritage Protection Unit of Carabinieri (Italian police corps), likely found in 1994 in Lazio or Campania; it retains the head similar to the one from Pompeii but with fewer details [17].

The four replicas were exhibited in 2023 at the MAAM in Grosseto in the exhibition "Una, Nessuna, Centomila,". In this context, despite being headless, the statue from Castiglion della Pescaia is confirmed to be perhaps the most refined in craftsmanship. The statue is in good condition and has not undergone significant interventions, aside from being mounted on a rotating pin fixed to its current stone base. These circumstances support further investigations, particularly considering the findings from the study of the replica found in Pompeii. The abundant traces of polychromy discovered on the head, chiton, and sandals of the Pompeii replica [14] encourage similar investigations and comparisons on the Castiglion della Pescaia version.

## 2 Methods

Several imaging methods have proved to be useful to investigate ancient artwork surfaces, as such we integrated into our 3D model investigation some of them to be used in the chosen classification method.

Imaging Methods: We have explored a combination of Ultraviolet-induced Fluorescence (UVF), Infrared reflectance (IRR), and visible light (VIS) imaging techniques. UVF is particularly effective in revealing organic materials such as binders and adhesives, providing insight into previous conservation interventions, while IRR can reveal complementary information on pigments and techniques.

Classification Method: The processing of multispectral and hyperspectral data relies on advanced classification algorithms to differentiate and identify the various materials and pigments present. In this study, we employ the Spectral Angle Mapper (SAM) algorithm. SAM is useful for identifying spectral similarities between known references and the acquired data, allowing for precise mapping of the spectral characteristics of polychrome remnants.

3D Digitization Method: To accurately map the spectral information in three-dimensional space, we integrate multispectral imaging with photogrammetry. The spectral data obtained from UVF and IRR imaging is then overlaid onto these 3D models, creating a comprehensive and spatially accurate representation of the residual polychromy. This integration facilitates a more nuanced understanding of the distribution and extent of the remaining pigments across the sculpture's surface.

In terms of symbology, we need to clarify the difference between "uv" and "UV" in the rest of this article. We use "uv" to refer to the computer graphics concept of a 2D coordinate system used to map textures onto the surface of 3D models. The names "u" and "v" indicate the horizontal and vertical axes of the 2D texture spaces, chosen arbitrarily to differentiate them from the x, y, and z coordinates used for the 3D model. Conversely, we use the uppercase "UV" as an abbreviation for ultraviolet.

### 2.1 Smartphone-based 3D Data acquisition

To provide a very accessible 3D model acquisition method, we experimented using a smartphone camera and the photogrammetry technique. The smartphone used for the data acquisition was a Samsung Galaxy A41, whose camera features a 48MP Sony IMX582 1/2.0" sensor with a 4.8mm f/2.0 lens, equivalent to a 26mm lens on a full-frame sensor. This sensor utilizes Quad Bayer color filtering and pixel binning, which means each pixel is generated by averaging the information from four neighboring pixels, resulting in 12MP photos from the captured 48MP data.

Given the fact that the statue could spin around its axis, we decided to shoot having a stationary camera mounted on a tripod, illuminating uniformly the volume in front of the camera, and rotating the statue. The angle of rotation between two consecutive images was 10 degrees. To control the rotation, a 360-degree protractor with 10-degree steps was drawn on black cardboard and attached to the stationary base of the sculpture with the center at the axis of rotation.

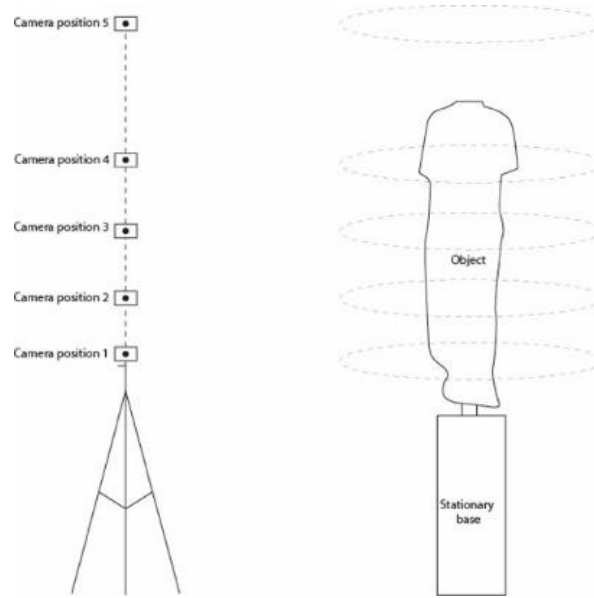

**Fig. 1.** Camera positioning around the sculpture to generate the photogrammetric images set.

Another pointer was attached to the rotating statue to control each rotation step. By using this method, we captured 36 images in total for a complete rotation of the object. Five such rotation sets were performed to cover the complete surface of the object starting from the base of the statue till the top as demonstrated in Fig. 1. Using the same protocol, the following three sets of images were acquired:
- Illumination: IR (incandescent lights); Filter: Infrared band pass
- Illumination: UV (Ultraviolet emitting lights); Filter: Ultraviolet band stop
- Illumination: white in the visible spectrum (LED lights); Filter: None

For IR and UVF acquisitions, the highest possible exposure time of 2 sec was applied by using the ISO values of 800 and 1000 respectively. For acquisition under visible light, a shorter exposure time of 1/30 sec was applied by using the ISO value of 200. The color temperature was fixed to 5000K for all acquisitions. The camera app was set to save the images in raw format, to take advantage of the 12-bit/channel sensor equipped on the A41 camera, avoiding the risk of clipping on color components due to saturation and allowing in this way a meaningful color correction of the images in post-processing. For scaling the models, the standard targets from Agisoft Metashape were used before each set of acquisitions.

### 2.2 Data processing to produce a multitextured 3D model

The photogrammetric process was carried out using Agisoft Metashape. Since the captured images were saved in RAW format as "Digital Negative" (DNG) files, we used the capability of that SW package to process DNG files, where the 12 bits of the information provided by the sensor for each channel are preserved in a 16-bit word.

Therefore, the alignment of the images was performed on the entire set of RAW images, including:
- All images taken with visible light, without filtering, including those with targets for photogrammetric scaling.
- All images taken with infrared lighting and captured in the very near infrared spectrum with a visible blocking filter (cutoff at 720 nm), excluding those with photogrammetric targets.
- All images taken with ultraviolet lighting and captured in the visible spectrum with a UV blocking filter, excluding those with photogrammetric targets.

In all sets, a reflectance calibration target remained visible. It is a 10 cm x 10 cm plate of Spectralon, a polymer exhibiting a highly Lambertian behavior and a reflectance higher than 99% over a range of wavelengths from 400 to 1500 nm.

Eventually, the image alignment successfully concluded with all the 573 available images (196 visible, 188 UVF, and 189 IRR) aligned in the same data set thanks to their wide dynamic range. This means that the sculpture acted as a reference for all images independently of the illumination source. Consequently, all images were oriented in space in the same reference system, making it possible to re-project the different sets on the same uv space.

The average reprojection error, which represents a quality index for both the camera calibration and the orientation of the images in space, was 0.6 pixels. A reprojection error below 1 pixel is generally considered good.

The tie points generated from this first phase were slightly over two hundred thousand (239,785), sufficient to visualize the sculpture's structure. This allowed the sculpture to be aligned with the Cartesian axes, with the z-axis in the vertical direction and the frontal face along the xz plane.

The sparse cloud of tie points also enabled the creation of a low-resolution pre-model of the sculpture, useful to create masks for the visible images. This operation is generally not necessary but represents the standard process we use in creating quality models of complex sculptures. Masking helps generate dense point clouds by minimizing the number of outliers and consequently reducing manual cleanup of the dense cloud before generating the final mesh.

The next step was scaling the model based on three distance measurements taken onsite with a metal tape measure. The average error on the control distances was 0.225 mm.

At this point, using the image-matching process, a dense cloud was created by selecting only the images originated by white light in the visible spectrum, which generated over 4 million 3D points (4,270,905). The cloud, already very clean thanks to the prior masking of the images, was further refined by eliminating points with low confidence (between 0 and 2), on a confidence scale determined by an 8-bit parameter (0 to 255).

The last phase of meshing involved creating a medium-resolution model (838,763 faces, 419,572 nodes), which was used as a basis for various texturing operations.

The initial texturing was achieved by exclusively selecting visible-spectrum images. During this phase, the Metashape system executed uv parameterization, mesh unwrapping, and image projection into the uv space. The chosen texture resolution was 8192

x 8192 pixels, i.e., 64 megapixels, stored as an EXR file (32 bit per channel) to retain the high dynamic range derived from the sensor (12 bit per channel). The original images, captured in RAW format with 12 bit per channel and saved as DNG files with 16 bits per channel, were processed within the photogrammetric pipeline using 16-bit mathematics before being merged into a 32-bit per channel texture, and eventually saved in EXR format.

With all images precisely aligned with respect to the sculpture, it was feasible to selectively use either UVF or IRR images to generate two additional textures utilizing the same texture parameterization. This process resulted in a trio of 8k x 8k square images representing the same uv space filled sequentially with visible, UVF, and IRR data, each corresponding pixel-by-pixel. Notably, the fact that the color values were represented using 32 bits per channel, has allowed the calibration directly in the uv space while preserving the original 12-bit per channel data originated by the sensor for each pixel.

However, it was not possible to further use the IRR images and the derived 3D model. As confirmed in the literature, the smartphone sensors are strongly filtered to cut IR components [18]. As a result, the mobile camera sensor's low sensitivity to those wavelengths required exposure times that were incompatible with the smartphone hardware. Even with the maximum exposure time allowed by the smartphone camera (30s), the resulting images were of insufficient quality to provide meaningful additional data.

An important outcome of the photogrammetric process involving images captured under different lighting conditions is worth specifically mentioning. These images, taken from slightly varied positions and angles, could not have been aligned individually; however, aligning them relatively to the sculpture's 3D shape enabled the projection into a consistent uv space, creating textures with exact pixel correspondence. Consequently, the 3D model of the sculpture features three alternative registered textures, allowing for pixel-level analysis of reflectance and emissions caused by different lighting conditions. This analysis aids subsequent classifications in mapping in 3D similar polychromy contents in different areas of the sculpture.

### 2.3    Images calibration

To get an illumination-independent rendering of both the visible and the UV fluorescence images, and ensure consistency in the reflectance and radiance levels, we had to apply an appropriate calibration to acquired images. Radiometric calibration of an image refers to the process of correcting the pixel values within the image to ensure accurate and consistent measurements of reflectance or radiance. This calibration is essential when precise quantitative analysis of image data is necessary. This step ensures that different images taken by the same sensor, or different sensors can be compared reliably, as they are adjusted to a common scale.

#### 2.3.1    Visible images calibration

During calibration, pixel values are often normalized using known reference targets or standard radiometric sources.

We performed the calibration for all acquired images after they had been included in the texture map. This was possible since all images in the visible dataset were acquired under very similar illumination conditions.

We used a reflectance standard (Spectralon, Labshere) that exhibits very high uniform reflectance over a wide range of wavelengths (from Ultraviolet to near-Infrared) and behaves as a near-perfect Lambertian reflector, meaning it reflects light uniformly in all directions. These characteristics ensure that the reflectance measurements are consistent and not dependent on the illumination angle or observation.

The intensities in the different RGB bands in the acquired images composing the texture map were normalized against the intensities registered on the Spectralon target.

From the median intensity values measured over a 10x10 pixel area of the Spectralon target, for each of the RGB components, $R_{target}$, we calculated the calibration vector $R_{norm}$ using the reflectance nominal value $R_{target}$ given by the manufacturer

$$R_{norm} = \frac{R_{target}}{R_{nominal}}$$

The $R_{norm}$ calibration vector was then used to convert the raw color pixels of the acquired texture map image to calibrated reflectance images (Fig. 2).

We can compute a calibrated image $V(x,y)_{calib}$ using for each of the 3 RGB bands, the image acquired in the visible range $V(x,y)_{acq}$

$$V(x,y)_{calib} = \frac{V(x,y)_{acq}}{R_{norm}}$$

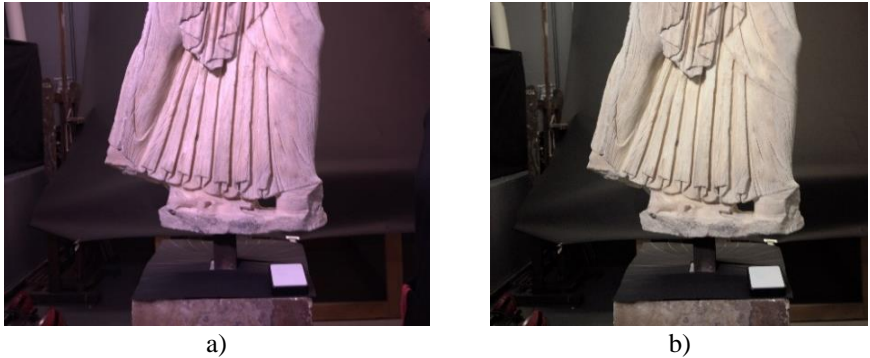

a) b)

**Fig. 2.** Example of image calibration: a) uncalibrated image; b) corresponding calibrated one.

### 2.3.2   UV Fluorescence images calibration

The UV fluorescence calibration process is meant to eliminate the contribution given by stray VIS light to the acquired UVF images.

Therefore, we first need to find the intensity of this VIS stray light, and then subtract it from each of the acquired UVF images.

Since the Spectralon reflectance standard does not emit any radiation due to fluorescence, the radiation observed on its surface when illuminated with UV light can only be due to a residual visible light emission of the UV lamps used, and/or other stray VIS light in the environment. We can then consider it as the total stray light.

Therefore, we considered the stray light $S_{target}$ as the median of the intensity values acquired over the 10x10 pixel area of the image of the Spectralon target, for each of the RGB components.

However, given the fact that the statue surface has a varying reflectance, to compute the value to be subtracted for each pixel of the UVF texture map, we had to consider not only the incident stray light but also the reflectance computed for each pixel in the VIS texture map.

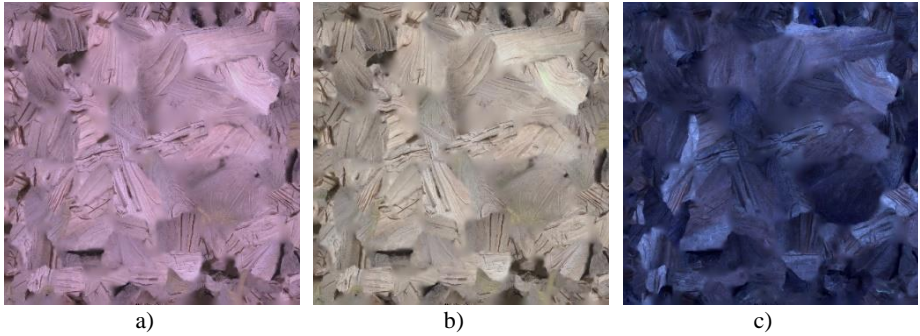

| a) | b) | c) |

**Fig. 3.** Texture maps originated by different calibrations and illuminations: a) uncalibrated visible; b) calibrated visible; c) calibrated UV.

We used this formula to compute the calibrated UVF image $U(x,y)_{calib}$ for each of the RGB components:
$$U(x,y)_{calib} = U(x,y)_{fluorescence} - S_{target} * V(x,y)_{calib}$$

Where $U(x,y)_{fluorescence}$ is the UVF uncalibrated texture map computed with the acquired UVF fluorescence images.

It is important to notice that since the visible reflectance texture map image is perfectly registered with the fluorescence texture map, the fluorescence calibration can be operated directly on the two aligned texture maps (Fig. 3).

### 2.4   Texture maps multispectral cube

The calibrated visible and fluorescence texture maps can now be used to obtain a multispectral representation of the statue surface by building on each pixel of the texture a vector with a calibrated value for each acquired waveband and imaging modality. We can thus produce a multispectral data cube, that is a three-dimensional dataset consisting of two spatial dimensions (x and y) and one spectral dimension (λ), where the spectral dimension captures the intensity of light at various wavelengths for and in different modalities. In this case λ =6, 3 dimensions for visible reflectance, corresponding to the VIS RGB bands and 3 dimensions for the UVF RGB image.

This same method could be applied using more images, acquired in more closely spaced wavelengths and/or more modalities and wavebands, for example, the Infrared band or the Visible Induced Florescence. This would increase the dimensions of the multispectral cube and allow for more accurate analysis and classification. In such a case, the methodology here presented for the analysis of the acquired data would deliver even more relevant results. This possibility is however linked to the availability of a suitable acquisition system. In the current work, we privileged the accessibility of the technique by choosing to capture images with the camera of a mobile phone (see above, paragraph 2.2, for a discussion on its limits).

### 2.5 Supervised material classification on the 3D model

The proposed classification method offers the possibility of spotting and highlighting directly on the 3D model, areas that are likely the same material, having similar spectral signatures, i.e. a unique pattern of electromagnetic radiation emitted, reflected, or absorbed across different wavelengths.

Given the fact that we have produced a multispectral texture maps cube, selecting a vertex of the 3D rendered model corresponds to a specific pixel on each of the registered textures.

In fact, each pixel in the texture maps has a corresponding spectral vector which is a vector of values, one for each of the sampled wavebands. A spectral vector has a close relation with a spectral signature.

To measure the similarity among vertex in the 3D model we used, we used the angle between spectral vectors in the texture maps multispectral cube.

The angle $\theta$ between two vectors **u** (the pixel's spectrum) and **v** (the reference spectrum) is calculated as follows:

$$\theta = cos^{-1}(\frac{\boldsymbol{u} \cdot \boldsymbol{v}}{\parallel \boldsymbol{u} \parallel \parallel \boldsymbol{v} \parallel})$$

Here, $\boldsymbol{u} \cdot \boldsymbol{v}$ is the dot product of the vectors, and $\parallel \boldsymbol{u} \parallel$ and $\parallel \boldsymbol{v} \parallel$ are the magnitudes of the vectors.

Two pixels are judged to be similar when the angle between their respective vectors is small enough, i.e., smaller than a predefined threshold.

Since this similarity measure is robust to changes in vector magnitude, it is effective in varying lighting conditions, thus making the analysis independent of a specific illumination [19].

Once we define a suitable threshold for $\theta$, using SAM classification algorithm [20], we can compare the spectral signature of the pixels in the multispectral cube to a reference spectra.

## 2.6 3D interface

When operators freely navigating the 3D model select a reference vertex, they can discover vertexes exhibiting similar spectral signatures, thus similar materials on the surface, wherever they are to be found on the acquired object.

By clicking on the 3D model (Fig. 4a), the corresponding reference pixel is located on the map (Fig 4b).

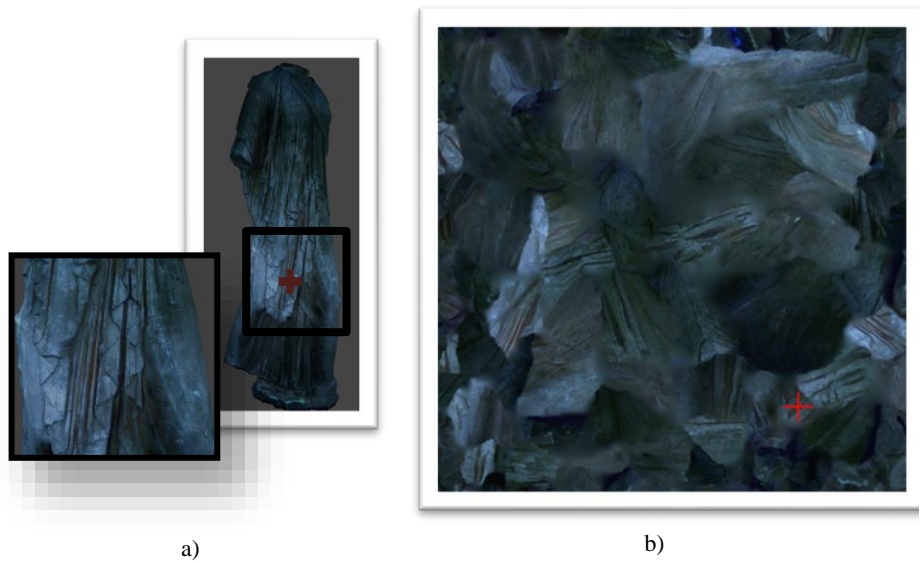

**Fig. 4.** Point of interest selection: a) a click on the 3D model surface identifies a point in 3D (black plus sign); b) corresponding point on the texture, represented in the uv space (red plus sign in the lower right corner).

Since the coordinates in the uv space are consistent across different color textures of the multispectral cube and the SAM map, these coordinates can be used to identify the corresponding element in the SAM map. This reference value can then be used to find other similar values within a certain threshold, which can subsequently be mapped back onto the 3D model.
In the example shown in Fig. 5 the set of points within an angular threshold of 0.15 radians from the point previously selected on the 3D model (Fig. 4), is identified on the SAM map (Fig. 5a), and then re-mapped on the 3D model (Fig. 5b). As a result, a region of points with spectral characteristics similar to those of the clicked point are immediately identifiable on the 3D surface of the sculpture.

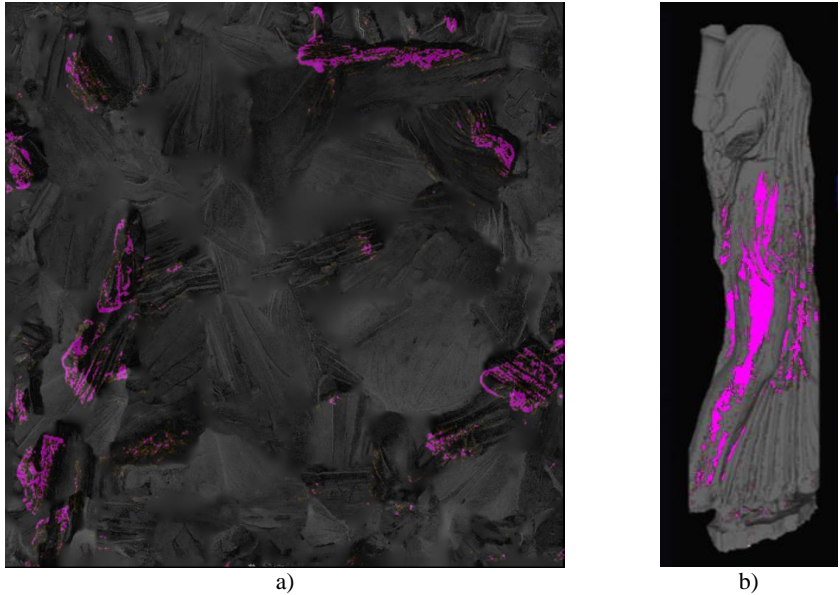

|  |  |
|:-:|:-:|
| a) | b) |

**Fig. 5.** Pixels with an angular deviation within 0.15 radians from the point selected in Fig. 4, highlighted in magenta on the SAM map represented on: a) the UV space; b) the 3D model.

## 3    Conclusions

The method presented in this paper offers a powerful tool for documenting the conservation status of ancient sculptures, including polychromy mapping. One notable achievement is the use of smartphone cameras to gather multispectral information, keeping operating costs relatively low.

This work demonstrates the ability to align images taken at different stages with various lighting sources and imaging modalities such as UVF, using the object itself as a spatial reference. By combining this feature with the high dynamic range provided by raw images, all images taken throughout the process can be aligned within a single reference system. Using a unified uv parameterization for all image sets enables the projection of different maps onto the same uv space with pixel-to-pixel correspondence. The result is a 3D map created from the integrated use of various uv maps. This allows for classification algorithms to identify areas with similar characteristics, which can reveal pigment residues, previous restoration materials, or similar conservation states. To this end we employed a customized supervised classification algorithm, using the SAM similarity measure on calibrated images. This innovative documentation method is valuable for planning restoration interventions.

However, a limitation of this method is the relatively low sensitivity of the camera sensor, necessitating longer exposure times. Therefore, using a rotating platform to rotate the sculpture is crucial, as it allows the camera to remain stationary on a tripod.

## 4    Acknowledments

We would like to thank the Soprintendenza Archeologia, Belle Arti e paesaggio per le province di Siena, Arezzo e Grosseto, and the Municipality of Grosseto - Museo Archeologico e d'Arte della Maremma for allowing us to have the statue of Artemis from Castiglion della Pescaia as a case study.